\documentclass{article}
\usepackage[preprint]{neurips_2025}

\usepackage[utf8]{inputenc}
\usepackage[T1]{fontenc}

\usepackage[dvipsnames,table]{xcolor}
\definecolor{linkColor}{RGB}{237,4,140}
\definecolor{citecolor}{RGB}{0, 113, 188}
\usepackage[colorlinks=true,linkcolor=linkColor,citecolor=citecolor,filecolor=linkColor,urlcolor=linkColor]{hyperref}       
\usepackage{amsmath}
\usepackage{amsfonts}
\usepackage{amssymb}

\usepackage{float}
\usepackage{xspace}
\usepackage{graphicx}
\usepackage{tabularx}
\usepackage{booktabs}
\usepackage{array}
\usepackage{makecell}
\usepackage{multirow}
\usepackage{mdframed}
\usepackage{tcolorbox}
\usepackage[hypcap=false]{caption}
\definecolor{customred}{RGB}{255, 99, 71}
\usepackage{tikz}
\usetikzlibrary{patterns}

\def\eg{\emph{e.g.}\xspace} 

\def\ie{\emph{i.e.}\xspace}

\definecolor{lightgray}{gray}{0.9} 
\definecolor{cvprblue}{rgb}{0.21,0.49,0.74}

\newcommand*{\system}{Seg-R1\@\xspace}

\makeatletter
\newcommand\figcaption{\def\@captype{figure}\caption}
\newcommand\tabcaption{\def\@captype{table}\caption}
\makeatother

\title{\system: Segmentation Can Be Surprisingly Simple with Reinforcement Learning}

\author{Zuyao You$^{1}$,~Zuxuan Wu$^{1}$\\
\\
$^{1}$Fudan University \\
\\
Project Page: \href{https://geshang777.github.io/seg-r1.github.io/}{https://geshang777.github.io/seg-r1.github.io}.
}

\begin{document}
\newcommand{\colorboxpattern}[1]{%
\begin{tikzpicture}
\fill[pattern=north east lines, pattern color=#1] (0,0) rectangle (0.6em,0.6em);
\draw[line width=0.5pt, color=#1] (0,0) rectangle (0.6em,0.6em);
\end{tikzpicture}%
}

\newcommand{\colordash}[1]{%
\begin{tikzpicture}
\draw[color=#1, line width=0.5pt, dash pattern=on 0.5pt off 0.5pt] (0,0) rectangle (0.6em,0.6em);
\end{tikzpicture}%
}
\maketitle

\begin{figure*}[!ht]
\centering
\vspace{-20pt}
\includegraphics[width=0.92\linewidth]{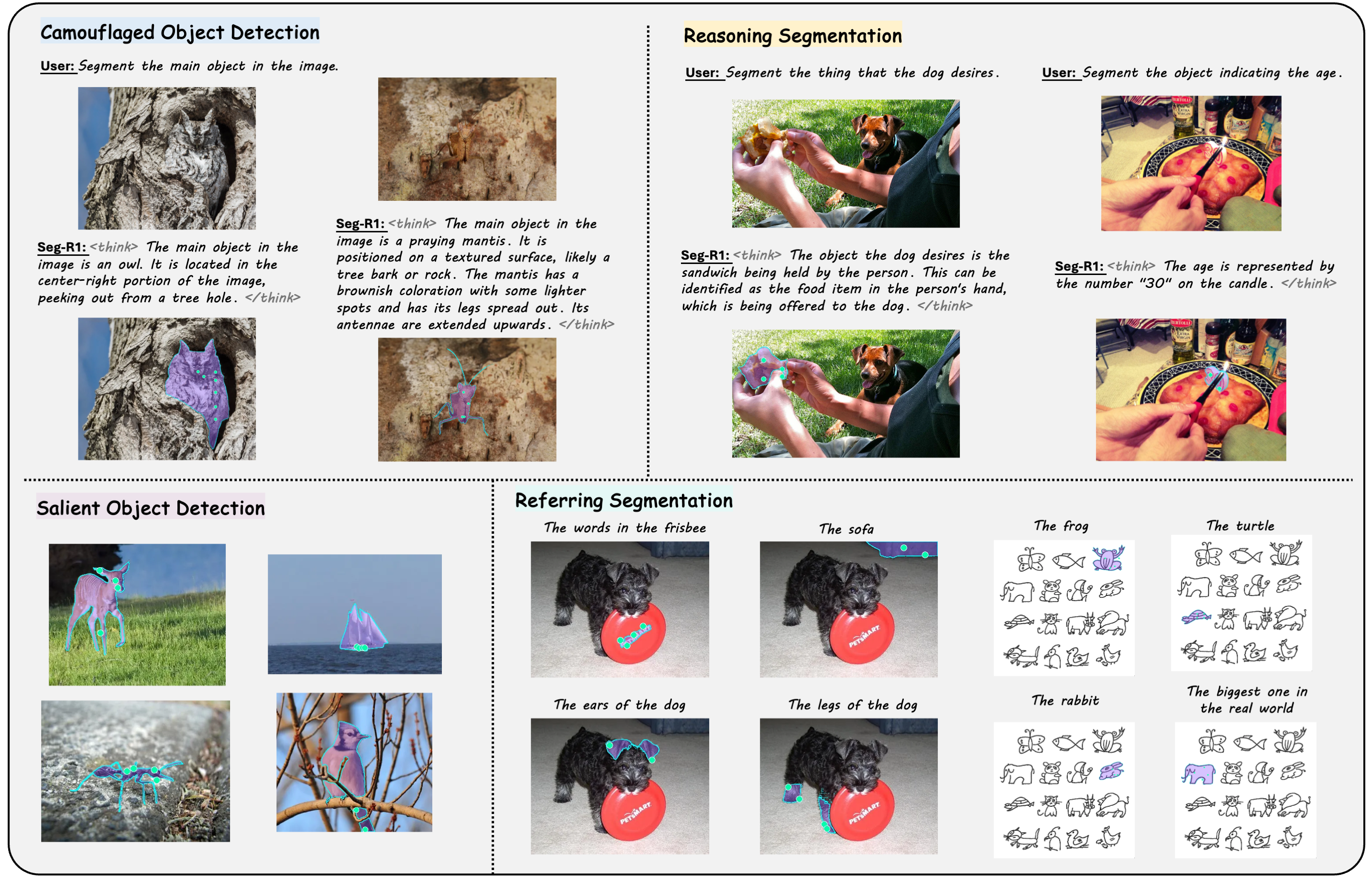}
\caption{Overview of segmentation ability of Seg-R1.}
\label{fig:teaser}
\end{figure*}


\begin{abstract}
We present \textbf{Seg-R1}, a preliminary exploration of using reinforcement learning (RL) to enhance the pixel-level understanding and reasoning capabilities of large multimodal models (LMMs). Starting with foreground segmentation tasks, specifically camouflaged object detection (COD) and salient object detection (SOD), our approach enables the LMM to generate point and bounding box prompts in the next-token fashion, which are then used to guide SAM2 in producing segmentation masks. We introduce \textbf{Group Relative Policy Optimization (GRPO)} into the segmentation domain, equipping the LMM with pixel-level comprehension through a carefully designed training strategy. Notably, Seg-R1 achieves remarkable performance with purely RL-based training, achieving \textbf{.873} S-measure on COD10K without complex model modification. Moreover,  we found that pure RL training demonstrates \textbf{strong open-world generalization}. Despite being trained solely on foreground segmentation image-mask pairs without text supervision, Seg-R1 achieves impressive zero-shot performance on referring segmentation and reasoning segmentation tasks, with \textbf{71.4} cIoU on RefCOCOg test and \textbf{56.7} gIoU on ReasonSeg test, outperforming models fully supervised on these datasets.




\end{abstract}

\section{Introduction}
\label{sec:intro}
Enhancing the granularity of image understanding in large multimodal models (LMMs) has long been a key challenge in the community. Early approaches have pushed the capabilities of image-level focused LMMs~\cite{li2022blip,li2023blip,openai2023gpt4,Qwen-VL,Qwen2-VL} towards region-level~\cite{liu2023visual,zhang2025gpt4roi,chen2023shikra,Qwen2.5-VL}. Recent research~\cite{lai2024lisa,zhang2024omg,ren2024pixellm,you2025pix2cap,yuan2025sa2va} has begun to explore ways to equip models with genuine pixel-level understanding. A common approach is to introduce special segmentation tokens into the model and design specialized decoder structures to decode these tokens into segmentation masks~\cite{lai2024lisa,zhang2024omg,yuan2025sa2va}. While effective to some extent, this method disrupts the continuity of the causal architecture. Moreover, training models for segmentation via supervised fine-tuning (SFT) requires large-scale datasets with pixel-level image-text annotations and extensive training time, limiting scalability~\cite{lai2024lisa,ren2024pixellm,zhang2024omg,yuan2025sa2va}.


Recently, reinforcement learning (RL)~\cite{mnih2015human,schulman2017proximal,rafailov2023direct,shao2024deepseekmath}, particularly Group Relative Policy Optimization (GRPO)~\cite{shao2024deepseekmath}, has been widely demonstrated as more effective compared to SFT. By optimizing over groups, GRPO enables the model to explore chain-of-thought (CoT) reasoning to solve complex problems and significantly reduces memory consumption during the RLHF~\cite{ouyang2022training} stage. Some recent efforts~\cite{chen2025r1v,shen2025vlm} have attempted to apply GRPO in the visual domain, proving its effectiveness in visual tasks like object counting. These studies confirm that RL, compared to SFT, can reach comparable results with substantially fewer training steps, suggesting a promising direction for efficient model training.

Inspired by these insights, we propose a new paradigm that leverages RL to equip LMMs with segmentation capabilities. We introduce \textbf{Seg-R1}, a simple yet effective framework for pixel-level learning. Our approach is built upon Qwen-2.5-VL~\cite{Qwen2.5-VL} and SAM2~\cite{ravi2024sam2}, where Qwen-2.5-VL is trained to generate bounding box and point prompts to guide SAM2 in producing segmentation masks. We incorporate \textbf{GRPO} into the segmentation task, requiring the model to output the reasoning process and mask prompts explicitly. To guide learning, we design a reward function that combines a format reward with a segmentation reward based on IoU and S-Measure~\cite{fan2017structure}, striking a balance between global accuracy and fine-grained structural fidelity.

To explore how far pure RL can drive segmentation in LMMs, we adopt a two-stage RL training strategy. Seg-R1 is first pre-trained with GRPO on the high-resolution DIS5K~\cite{qin2022} dataset to acquire fundamental knowledge of segmentation structure and formatting. It is then further fine-tuned on COD10K~\cite{fan2020camouflaged} and CAMO~\cite{le2019anabranch} to enhance both its segmentation precision and reasoning ability. Notably, our method requires no architectural modifications to Qwen-2.5-VL and introduces no special tokens. Seg-R1 autonomously learns to construct annotation trajectories and generate high-quality prompts for SAM2. As a result, it achieves remarkable performance on camouflaged object detection tasks, with an $S_{\alpha}$ of \(.873\) on COD10K-Test and \(0.826\) on CAMO. Besides, with further fine-tuning on DUTS~\cite{wang2017learning}, it achieves state-of-the-art (SoTA) performance on salient object detection, reaching an $S_{\alpha}$ of \(.878\) on DUT-OMRON~\cite{yang2013saliency}.

To further compare the effectiveness of RL with supervised fine-tuning (SFT), we introduce the \textbf{Foreground Chain-of-Thought (FCoT)} dataset to SFT the model as a cold start. FCoT is designed to replicate the step-by-step reasoning process a human annotator follows when using SAM2 to generate masks. It comprises \(1,500\) image–mask pairs collected from existing foreground segmentation datasets. Each pair was re-annotated by fitting the foreground masks using SAM2, guided by carefully constructed bounding boxes and point prompts in accordance with a standardized annotation protocol. To explicitly capture the annotators' reasoning process, we further leverage Gemini-2.5-Pro~\cite{gemini25report} to generate a natural language chain-of-thought of the annotation steps based on the provided prompts and annotation rules. We use FCoT to fine-tune Qwen-2.5-VL via supervised learning, serving as a comparative baseline against the RL-trained version of our model.


An additional interesting finding is that, as mentioned earlier, Seg-R1 is trained solely on \(7,040\) foreground segmentation image-mask pairs without any textual supervision and without exposure to referring segmentation and reasoning segmentation tasks. Despite this, it demonstrates remarkable open-world segmentation capabilities. In zero-shot evaluations on the RefCOCO~\cite{kazemzadeh2014referitgame}, RefCOCO+~\cite{yu2016modeling}, RefCOCOg~\cite{yu2016modeling}, and ReasonSeg~\cite{lai2024lisa} benchmarks, Seg-R1 achieves performance comparable to models trained directly on these datasets. Specifically, Seg-R1 attains a cIoU of \(71.4\)  on the RefCOCOg test and \(56.7\) gIoU on the ReasonSeg test.

Moreover, we observe that pure RL training for segmentation does not compromise the general-purpose capabilities of the model. On general multimodal benchmarks such as MMBench~\cite{liu2024mmbench}, MME~\cite{fu2023mme}, POPE~\cite{li2023evaluating}, and AI2D~\cite{kembhavi2016diagram}, Seg-R1 maintains performance on par with the original Qwen-2.5-VL. In contrast, the model fine-tuned via SFT experiences a noticeable performance drop. This highlights the efficacy of RL in enhancing pixel-level abilities without eroding the existing strengths of LMMs.

Our contributions can be summarized as follows:

\begin{itemize}
    \item We propose Seg-R1, a simple yet effective RL-based framework for enabling pixel-level segmentation in LMM.
    \item We introduce the FCoT, comprising \(1,500\) manually annotated mask prompts, which provides a valuable resource for prompt-guided segmentation.
    \item Through comprehensive experiments, we demonstrate that pure RL training equips LMM with strong segmentation capabilities while preserving their original visual comprehension ability, outperforming SFT in terms of generalization and efficiency.
\end{itemize}

\section{Related Work}

\paragraph{Large Multimodal Models for Pixel-Level Understanding.} 
Recent advances in LMMs have significantly enhanced the granularity of visual understanding, progressing from image-level~\cite{li2022blip,li2023blip,openai2023gpt4,Qwen-VL,Qwen2-VL} to pixel-level comprehension~\cite{lai2024lisa,zhang2024omg,ren2024pixellm,you2025pix2cap,yuan2025sa2va}. LISA~\cite{lai2024lisa} first introduced the \texttt{<SEG>} token into LMMs, enabling LMMs with reasoning segmentation ability. Following this idea, incorporating specialized segmentation tokens to endow LMMs with pixel-level capabilities~\cite{zhang2024omg,rasheed2024glamm,zhang2025pixel} has become a paradigm. For example, GLaMM~\cite{rasheed2024glamm} extends the \texttt{<SEG>} token into the multi-object scenario, enabling pixel-level grounded conversation generation. In a similar vein, OMG-LLaVA deploys a universal visual encoder, supporting a wide range of visual and textual prompts within a unified framework. Despite these advances, the mixture of text tokens and segmentation tokens may lead to confusion in next-token prediction. Furthermore, SFT on pixel-level tasks can often lead to catastrophic forgetting of general-purpose capabilities~\cite{lai2024lisa,ren2024pixellm}. In contrast, we propose a novel approach that enables LMMs to perform pixel-level segmentation without complex architectural modifications, preserving both simplicity and effectiveness.

\paragraph{Reinforcement Learning for Large Multimodal Models.} 

Reinforcement learning (RL)~\cite{sutton1998reinforcement} has emerged as a powerful approach to enhance the training of large multimodal models (LMMs). Among the various RL techniques~\cite{mnih2015human,schulman2017proximal,rafailov2023direct,shao2024deepseekmath}, Group Relative Policy Optimization (GRPO)~\cite{shao2024deepseekmath} has shown particular promise. Unlike traditional RL methods, which rely on a critic model to estimate the baseline, GRPO estimates the baseline from group scores, significantly reducing the computational resources required during training. This reduction in resource consumption has made GRPO an appealing option for training large-scale models. Recent works~\cite{chen2025r1v,shen2025vlm,wang2025simplear} have demonstrated the effectiveness of incorporating GRPO into LMMs, yielding substantial improvements in both visual generation and understanding tasks. In this work, we introduce GRPO to the pixel-level tasks, providing a more effective approach for improving the performance of LMMs in tasks requiring pixel-level precision.


\section{Methods}
\subsection{Seg-R1}
Segmentation has long been a challenging task in computer vision~\cite{cheng2021per,cheng2021mask2former,kirillov2019panoptic}. Training models to produce accurate masks typically requires large amounts of manually annotated pixel-level data~\cite{lin2014microsoft,plummer2015flickr30k,zhou2017scene}. Furthermore, generating fine-grained segmentation masks usually relies on training dedicated decoder architectures that transform visual features into dense outputs~\cite{cheng2021per,cheng2021mask2former,zou2023generalized}, which imposes significant computational costs.


In this work, we propose a much more efficient paradigm for segmentation. We employ Qwen-2.5-VL~\cite{Qwen2.5-VL} to predict points, bounding boxes, and labels (referred to as mask prompts) to guide SAM2~\cite{ravi2024sam2} in mask generation. This approach reduces the dense prediction of segmentation to a sparse mask prompting task, significantly lowering the learning cost. Besides, using a causal LLM to predict mask prompts aligns naturally with how human annotators think and then generate masks step by step, mirroring the autoregressive nature of next-token prediction. The conditional probability of each mask prompt \( m_t \) can be formulated as \(P(m_t \mid I,r, m_{<t})\),  where \( m_t \) is predicted based on the input image \( I \), the reasoning process \(r\) and the sequence of previously prompts \( m_{<t} \). 




\begin{figure*}[t]
\centering
\includegraphics[width=\columnwidth]{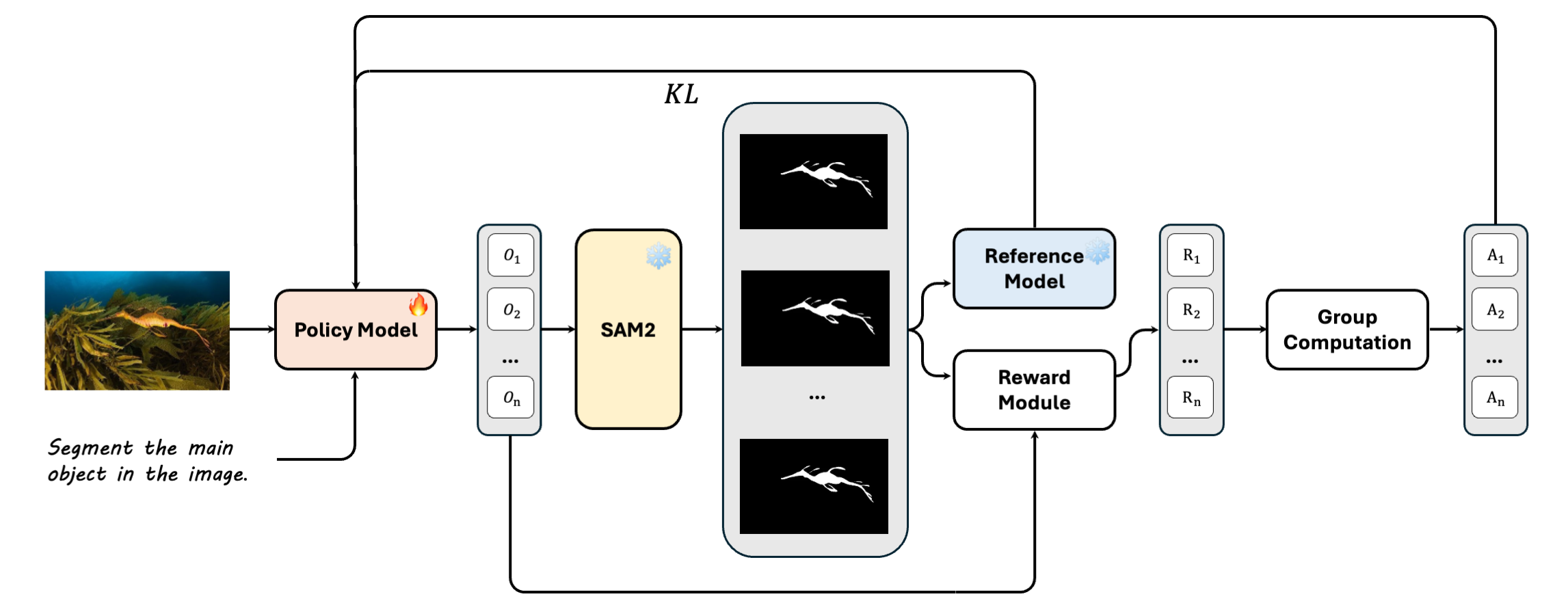}
\caption{Overview of the Seg-R1 framework. We introduce GRPO into the segmentation domain, enabling the model to develop pixel-level understanding through group-based advantage optimization, achieving strong segmentation performance without complex architectural modifications.}
\label{fig:framework}
\end{figure*}




We introduce GRPO into the segmentation task to enable effective reinforcement learning for the mask prompts. As illustrated in Figure~\ref{fig:framework}, Given an input image and a query \( q \), GRPO samples a group of outputs sequences \( o_1, o_2, \ldots, o_G \) from the current policy \( \pi_{\theta_{\text{old}}} \), and updates the policy \( \pi_{\theta} \) by maximizing the following objective:

\begin{equation}
    \mathcal{J}_{GRPO}(\theta) = \mathbb{E}_{o_{i} \sim \pi_{\theta_{old}}} \left[
    \begin{aligned}
        &\frac{1}{G}\sum_{i=1}^{G} \mathrm{min} \left( \frac{\pi_{\theta}(o_{i} | t)}{\pi_{\theta_{old}}(o_{i} | t)} \hat{A}_{i}, \mathrm{clip} \left( \frac{\pi_{\theta}(o_{i} | t)}{\pi_{\theta_{old}}(o_{i} | t)}, 1 - \epsilon, 1 + \epsilon \right) \hat{A}_{i} \right) \\
        &\hspace{5cm} -\beta \mathrm{D}_{KL} (\pi_{\theta} || \pi_{ref})
    \end{aligned}
    \right]
\end{equation}

Here, \( \hat{A}_{i} \) represents the estimated advantage, which measures the relative quality of an output within a group. The policy update is clipped by a threshold $\epsilon$ to ensure stable training. To prevent the updated policy from drifting too far from the original behavior, a KL divergence penalty is applied between the \( \pi_{\theta} \) and the reference model \( \pi_{\text{ref}} \), scaled by a hyper-parameter $\beta$.


The reward module is a combination of the format reward and the segmentation reward. The format reward is assigned a value of \(1.0\) if the output strictly wraps the generated content within the designated tags (\texttt{\(<\)think\(>\)}\texttt{\(<\)/think\(>\)}, \texttt{\(<\)bbox\(>\)}\texttt{\(<\)/bbox\(>\)}, \texttt{\(<\)points\(>\)}\texttt{\(<\)/points\(>\)}, and \texttt{\(<\)labels\(>\)}\texttt{\(<\)/labels\(>\)}); otherwise, the format reward is \(0\). We initially use S-Measure~\cite{fan2017structure} as the segmentation reward. S-measure combines both structural and content similarity between the predicted and ground truth, better aligns with human perceptual understanding by capturing both global structure and local consistency. However, using S-measure alone leads to reward hacking as mentioned in Sec~\ref{sec:ablation}. To mitigate this, we use a weighted combination of IoU (\(0.7\)) and S-measure (\(0.3\)) as the segmentation reward, promoting both global accuracy and structural consistency.


\subsection{FCoT}

To further investigate the differences between supervised fine-tuning (SFT) and reinforcement learning (RL) in the segmentation domain, we require a dataset that supports structured prompt-based supervision, specifically, one that incorporates bounding boxes and point-based prompts to guide SAM2 in mask generation. However, such datasets are currently lacking. To fill this gap, we introduce \textbf{FCoT}, a dataset specifically designed to capture the step-by-step reasoning process that human annotators follow when using SAM2 to produce segmentation masks.

\begin{figure*}[t]
\centering
\includegraphics[width=\columnwidth]{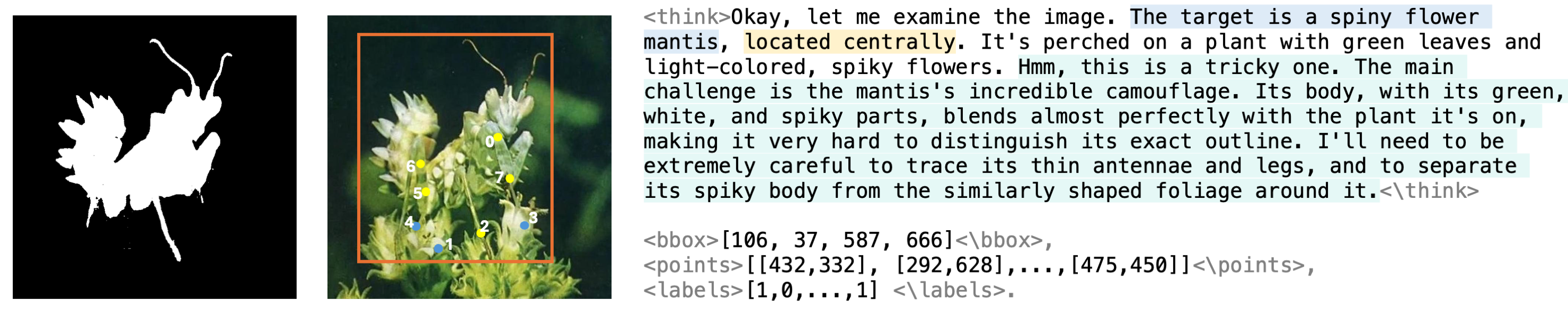}
\caption{A visualization of FCoT, a dataset that captures the step-by-step reasoning process used by human annotators to guide SAM2 in generating segmentation masks.}
\label{fig:fcot_vis}
\end{figure*}

In practice, human annotators interacting with SAM2 follow a standardized multi-step procedure: (1) identifying and locating the target object, (2) grounding the object using a bounding box, and (3) refining the segmentation using a combination of foreground and background points. During the annotation process for FCoT, we adhered strictly to this protocol and carefully recorded all points and bounding box prompts. This enables the model to explicitly learn from the reasoning trajectory embedded in human annotation behavior. To enrich the dataset with interpretable reasoning paths, we utilized Gemini-2.5-Pro~\cite{gemini25report} to generate natural language explanations based on the image content, recorded annotation steps, and the annotation rules.

FCoT comprises 1,500 image–mask pairs curated from existing foreground segmentation datasets: \(1,000\) images from DUTS~\cite{wang2017learning}, \(400\) from COD10K~\cite{fan2020camouflaged}, and \(100\) from CAMO~\cite{le2019anabranch}. Each mask was re-annotated by replacing the original dense mask with structured mask prompt sequences and corresponding chain-of-thought annotations. We provide a visualization of FCoT in Figure~\ref{fig:fcot_vis}.



\section{Experiments}

\subsection{Training Strategies and Implementation Details}

We explore two training paradigms for Seg-R1: SFT followed by RL, and pure RL from scratch.

\noindent\textbf{SFT Followed by RL.}
To provide Seg-R1 with a solid initialization, we perform supervised fine-tuning (SFT) on the FCoT for one epoch as a cold start. This step equips the model with a basic understanding of output format and prompt-guided segmentation. During SFT, we use a learning rate of \(2\times10^{-5}\) and a batch size of \(128\). Following this stage, we apply reinforcement learning using the GRPO~\cite{shao2024deepseekmath} algorithm. The model is trained exclusively on the camouflaged object detection dataset COD10K~\cite{fan2020camouflaged} and CAMO~\cite{le2019anabranch}, with four samples generated per prompt during policy optimization. To improve efficiency and memory usage, we employ vLLM~\cite{kwon2023efficient} with Flash Attention V2~\cite{dao2023flashattention}. During the RL stage, the learning rate is set to \(10^{-6}\) with a batch size of \(24\).

\noindent\textbf{Pure RL Training.}
To investigate the full potential of reinforcement learning in pixel-level tasks, we also train Seg-R1 entirely from scratch using RL. We begin with pre-RL training on the DIS5K-TR~\cite{qin2022} dataset, which consists of \(3,000\) high-resolution images with meticulously annotated masks. Compared to low-resolution datasets, DIS5K provides richer structural details, encouraging more nuanced model reasoning. We only require the model to generate points and labels as mask prompts in this stage. After pre-RL, we further RL the model on COD10K and CAMO. We use the same RL recipe as in the previous paradigm. Images are resized to \(768\times768\) for training and \(1024\times1024\) for inference. All experiments are conducted on 8 NVIDIA A100 GPUs with 80G memory.

\subsection{Benchmark and Metrics}
\noindent\textbf{Foreground segmentation.} We evaluate Seg-R1 on two representative tasks: camouflaged object detection (COD) and salient object detection (SOD). For COD, we adopt widely used CAMO~\cite{le2019anabranch} and COD10K~\cite{fan2020camouflaged} as benchmarks. And for SOD, we use DUTS-TE~\cite{wang2017learning}, DUT-OMRON~\cite{yang2013saliency}, HKU-IS~\cite{li2015visual}, and ECSSD~\cite{shi2015hierarchical} as benchmarks. We use S-measure ($S_{\alpha}$)~\cite{fan2017structure}, E-measure ($E_{\phi}$)~\cite{fan2018enhanced}, F-measure ($F_{\beta}$) (weighted F-Measure for COD and max F-Measure for SOD), and Mean Absolute Error ($M$) as metrics, which are commonly used in foreground segmentation to comprehensively evaluate the models.

\noindent\textbf{Referring segmentation and reasoning segmentation.} For referring segmentation, we evaluate on RefCOCO~\cite{kazemzadeh2014referitgame}, RefCOCO+~\cite{yu2016modeling}, and RefCOCOg~\cite{yu2016modeling}, using cIoU as the metric. For reasoning segmentation, we conduct experiments on the ReasonSeg~\cite{lai2024lisa} dataset and report both cIoU and gIoU to capture the capability of the model in open-world pixel-level understanding.

\begin{table*}[t]
    \centering
    \setlength{\tabcolsep}{15pt}
    \caption{Results on Camouflaged Object Detection (COD). $\diamond$ indicates SFT on FCoT as a cold start.}
    \label{tab:cod}
    \renewcommand{\arraystretch}{1.2}
    \resizebox{0.95\textwidth}{!}{

    \begin{tabular}{c|cccc|cccc}
        \hline
        \multirow{3}{*}{} 
        & \multicolumn{4}{c|}{COD10K~\cite{fan2020camouflaged}} 
        & \multicolumn{4}{c}{CAMO~\cite{le2019anabranch}} \\
        \cline{2-9}
        & \small{$S_{\alpha}\uparrow$} & \small{$F_{\beta}\uparrow$} & \small{$E_{\phi}\uparrow$} & \small{$M\downarrow$}
        & \small{$S_{\alpha}\uparrow$} & \small{$F_{\beta}\uparrow$} & \small{$E_{\phi}\uparrow$} & \small{$M\downarrow$} \\
\hline
\multicolumn{9}{c}{Mask Supervision Setting} \\\hline
PFNet~\cite{mei2021camouflaged}    & .800 & .660 & .868 & .040 & .782 & .695 & .852 & .085 \\
ZoomNet~\cite{pang2022zoom}    & .838 & .729 & .911 & .029 & .820 & .752 & .892 & .066 \\
BSA-Net~\cite{zhu2022can}    & .818 & .699 & .901 & .034 & .794 & .717 & .867 & .079 \\
FSPNet~\cite{huang2023feature}    & .851 & .735 & .895 & .026 & .856 & .799 & .899 & .050 \\
ZoomNeXT~\cite{pang2024zoomnext}  & .898 & .827 & .956 & .018 & .889 & .857 & .945 & .041 \\
BiRefNet~\cite{zheng2024bilateral}  & \textbf{.912} & .874 & .960  & .014 & .904 & .890 & .954 & .030 \\
FOCUS~\cite{you2025focus} &.910 &\textbf{.883} &\textbf{.974} &\textbf{.013} &\textbf{.912} &\textbf{.904} &\textbf{.963} &\textbf{.025} \\
        \hline

\multicolumn{9}{c}{Prompt-Guided Setting} \\
\cline{1-9}
SAM\cite{kirillov2023segment} & .730 & .673 & .737 & .093 & .643 & .597 & .639 & .160 \\
GPT4V+SAM \cite{openai2023gpt4,kirillov2023segment}& .601 & .448 &.672 & .187 & .618 & .613 & .655 & .265 \\
GenSAM~\cite{hu2024relax} & .783 & .695 & .843 & .058 & .729 & .669 & .798 & .106 \\
ProMaC~\cite{hu2024leveraging} & .805 & .716 & .876 & .042  & .767 & .725 & .846 & .090 \\
Grounded SAM2~\cite{ren2024grounded} & .686 & .628 & .722 & .209 & .578 & .571 & .612 & .302 \\

\rowcolor{lightgray}\textbf{Seg-R1-3B}$\diamond$ & .850 & .798 & .902 & .036 & .810 & \textbf{.798} & .861 & .077 \\ 
\rowcolor{lightgray}\textbf{Seg-R1-3B} 	& .857 & .816 & .908 & .033 & .805 & .797 & .853 & .079 \\
\rowcolor{lightgray}\textbf{Seg-R1-7B} 	& \textbf{.873} & \textbf{.820} & \textbf{.926} & \textbf{.031} & \textbf{.826} & .788 & \textbf{.881} & \textbf{.073} \\
\hline
\end{tabular}

}

\end{table*}

\begin{table*}[t]
    \centering
    \setlength{\tabcolsep}{4pt}
    \caption{Results on Salient Object Detection (SOD). $\diamond$ indicates SFT on FCoT as a cold start, \textbf{ft} refers the version fine-tune on the DUTS-TR.}
    \label{tab:sod}
    \renewcommand{\arraystretch}{1.3}
    \resizebox{1.0\textwidth}{!}{
    \begin{tabular}{c|cccc|cccc|cccc|cccc}
        \hline
        \multirow{3}{*}{} 
        & \multicolumn{4}{c|}{DUT-OMRON~\cite{yang2013saliency}} 
        & \multicolumn{4}{c|}{DUTS-TE~\cite{wang2017learning}} 
        & \multicolumn{4}{c|}{HKU-IS~\cite{li2015visual}} 
        & \multicolumn{4}{c}{ECSSD~\cite{shi2015hierarchical}} \\
        \cline{2-17}
        & \small{$S_{\alpha}\uparrow$} & \small{$F_{\beta}\uparrow$} & \small{$E_{\phi}\uparrow$} & \small{$M\downarrow$} 
        & \small{$S_{\alpha}\uparrow$} & \small{$F_{\beta}\uparrow$} & \small{$E_{\phi}\uparrow$} & \small{$M\downarrow$} 
        & \small{$S_{\alpha}\uparrow$} & \small{$F_{\beta}\uparrow$} & \small{$E_{\phi}\uparrow$} & \small{$M\downarrow$} 
        & \small{$S_{\alpha}\uparrow$} & \small{$F_{\beta}\uparrow$} & \small{$E_{\phi}\uparrow$} & \small{$M\downarrow$} \\
\hline
\multicolumn{17}{c}{Mask Supervision Setting} \\\hline

        BBRF~\cite{ma2021receptive}     &.855 &.843 &.887  &\textbf{.042} &.908 &.916 &.927 &.025 & \textbf{.935} &\textbf{.958} &.965  &.020 &.939 &.963 &.934  &.022\\
        EVPv1~\cite{liu2023explicit}    &.862 &\textbf{.858} &.894  &.046  &.913  &.923 &.947 &.026 &.931 &.952 &.961  &.024  &.935 &.960 &.957  &.027 \\
        EVPv2~\cite{liu2023explicit2}     &.862 & .857 &.895 &.047 &.915 &.923 &.948 &.027  &.932 &.953 &.963  &.023  &.935 &.958 &.957  &.028  \\

        SelfReformer~\cite{10287608}     &.859	&.838 &.884	&.043 &.911 &.916 &.920 &.026 &.930 &.947 &.959 &.024 &.941 &\textbf{.963} &.935 &.025	 \\
        FOCUS~\cite{you2025focus} &\textbf{.868} &.836 &\textbf{.900} &.045 &\textbf{.929} &\textbf{.928} &\textbf{.965} &\textbf{.019} &\textbf{.935} &.942 &\textbf{.974} &\textbf{.018} & \textbf{.943} &.954 &\textbf{.971} &\textbf{.018} \\
        \hline
\multicolumn{17}{c}{Weakly Supervised Setting} \\\hline

NSAL~\cite{piao2022noise} & .745 & .648 & .801 & .088 & .782 & .730 & .849 & .073 & .854 & .864 & .923 & .051 & .834 & .856 & .883 & .078 \\
SLF~\cite{liu2023novel}   & .778 & .708 & .844 & .066 & .793 & .750 & .865 & .060 & .851 & .857 & .923 & .050 & .864 & .878 & .908 & .056 \\
HSS~\cite{cong2022weakly} & --   & --   & --   & .050 & .837 & .807 & --   & .050 & .887 & .892 & --   & .038 & .886 & .899 & --   & .051 \\
A2S~\cite{zhou2023texture} & .719 & .841 & .069 & --   & .750 & .860 & .065 & --   & .887 & .937 & .042 & --   & .888 & .911 & .064 & -- \\
A2SV2~\cite{zhou2022activation} & -- & .731 & .851 & .065 & -- & .767 & .871 & .061 & -- & .891 & .939 & .041 & -- & .902 & .923 & .056 \\
A2SV3~\cite{yuan2024unified}   & -- & .759 & .868 & .062 & -- & .816 & .906 & .047 & -- & .908 & .954 & .033 & -- & .923 & .951 & .038 \\

        \hline
\multicolumn{17}{c}{Prompt-Guided Setting} \\\hline
Grounded SAM2~\cite{ren2024grounded} &.708&.590&.744&.102 &.800 &.748 &.833 &.078 &.825&.842&.864&.069 &.845 &.849 &.871 &.078\\

\rowcolor{lightgray}\textbf{Seg-R1-3B}$\diamond$ 
& .837 &.789 & .878 & .048 & .861 & .848 & .856 & .048 & .780 & .818 & .820 & .077 & .881 & .871 & .901 & .052 \\

\rowcolor{lightgray}\textbf{Seg-R1-3B} 
& .852 &.821 & .890 & .045 & .866 & .863 & .899 & .045 & .772 & .814 & .809 & .078 & .882 & .908 & .903 & .050 \\

\rowcolor{lightgray}\textbf{Seg-R1-3B (ft)} 
& .868 & .839 & .906 & .047 & .909 & .914 & .942 & .031 & .890 & .914 & .925 & .041 & .916 & .939 & .940 & .036 \\

\rowcolor{lightgray}\textbf{Seg-R1-7B (ft)} 
& \textbf{.878} & \textbf{.850} & \textbf{.911} & \textbf{.045} & \textbf{.925} & \textbf{.922} & \textbf{.953} & \textbf{.025} & \textbf{.935} & \textbf{.950} & \textbf{.966} & \textbf{.022} & \textbf{.939} & \textbf{.956} & \textbf{.962} & \textbf{.025} \\
        \hline
    \end{tabular}
    }
\end{table*}

\subsection{Comparing with State-of-the-arts}
\noindent\textbf{Results on COD.} COD focuses on segmenting disguised objects that blend seamlessly into their surroundings, \eg, mimetic organisms. This task is particularly challenging due to the low visual saliency and ambiguous object boundaries. We compare the recently proposed methods that directly supervise the mask or supervise prompts (\ie, points, bounding box, and scribble) in COD. As shown in the Table~\ref{tab:cod}, Seg-R1 consistently outperforms the prompt supervised methods, surpassing strong baselines such as Grounding SAM2~\cite{ren2024grounded}, ProMaC~\cite{hu2025int}, and GenSAM~\cite{hu2024relax} by a significant margin. But, considering that some of these prompt-guided methods are weakly supervised, we also compare Seg-R1 with previous fully supervised methods. However, the segmentation capability of Seg-R1 relies heavily on frozen SAM2, which is known to struggle with camouflaged objects~\cite{hu2024relax}. As a result, there remains a performance gap on CAMO between Seg-R1 and prior fully supervised models specifically designed for camouflage segmentation. Nevertheless, our model still achieves competitive scores on COD10K, with an $S_{\alpha}$ of \(.873\) on COD10K-Test.

\begin{figure*}[t]
\centering
\includegraphics[width=1.0\columnwidth]{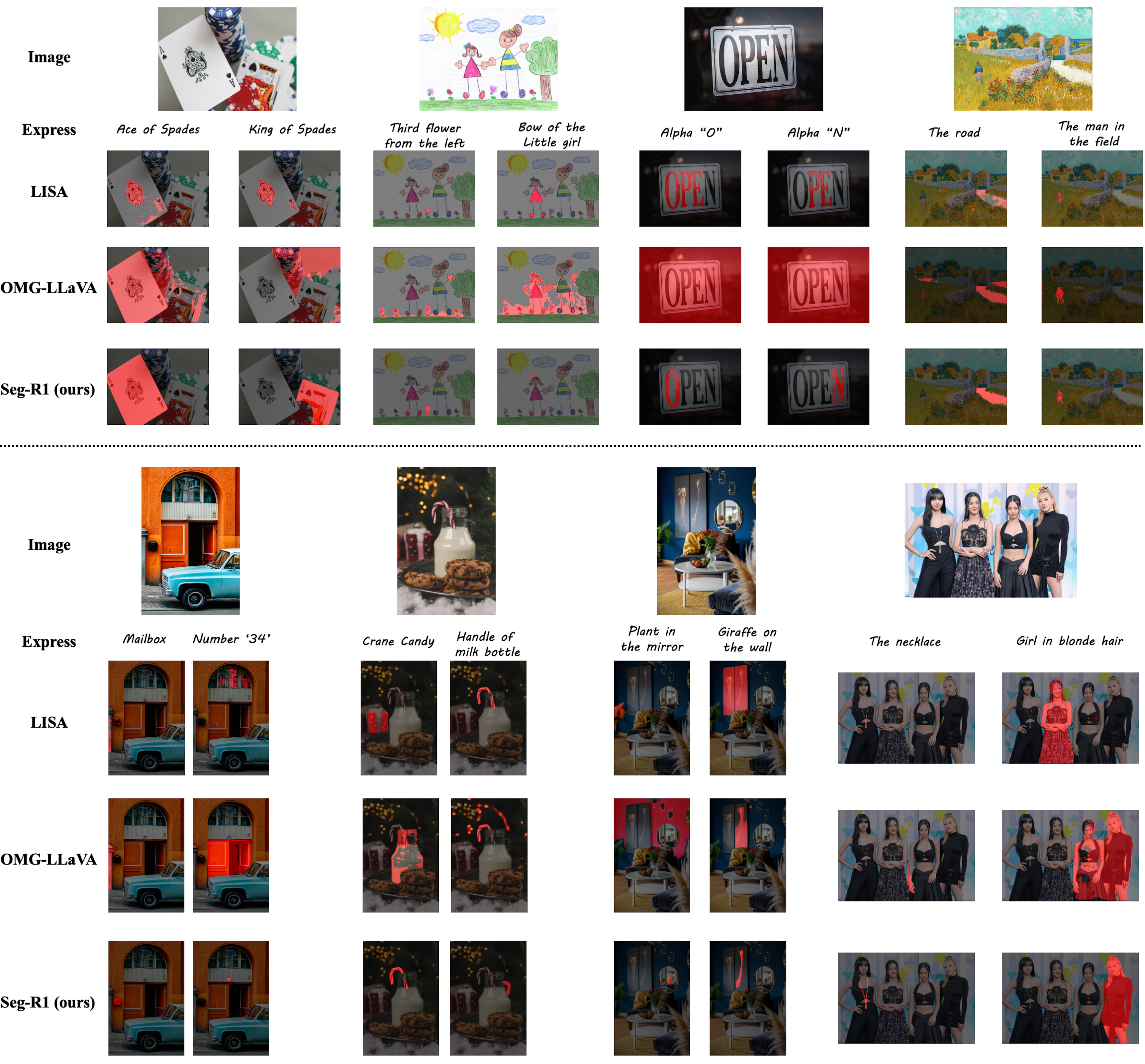}
\caption{Comparison of single object referring segmentation in the wild.}
\label{fig:comparison}
\end{figure*}

\noindent\textbf{Results on SOD.} SOD aims to segment the most salient part of the image from the background. We compare the recent SOD methods on four widely used benchmarks in the Table~\ref{tab:sod}. We first evaluate the zero-shot performance on SOD, with the version RL only on the DIS5K, COD10K, and CAMO datasets. As shown in the table, Seg-R1 demonstrates comparable performance with previous SoTA, which is supervised on DUTS datasets. We further fine-tune Seg-R1 on the training set of DUTS and find that Seg-R1 achieves SoTA performance in the salient object detection benchmarks. Notably, Seg-R1 achieves an $S_{\alpha}$ of \(.878\) and an $E_{\phi}$ of \(.911\) on DUT-OMRON~\cite{yang2013saliency}, outperforming the previous mask or prompt supervised methods with a clear margin.


\begin{table*}[t]
\centering
\begin{minipage}[t]{0.57\linewidth}
\centering
\setlength{\tabcolsep}{1.0mm}
\scriptsize
\caption{Results on referring segmentation.}
\vspace{0.05in}
\begin{tabular}{c*{3}{c}*{3}{c}*{2}{c}}

\toprule
& \multicolumn{3}{c}{RefCOCO} 
& \multicolumn{3}{c}{RefCOCO+} 
& \multicolumn{2}{c}{RefCOCOg}  \\
\cmidrule(lr){2-4}\cmidrule(lr){5-7}\cmidrule(lr){8-9}
& testA  & testB & val  
& testA  & testB & val  
& test & val\\
\midrule
LAVT~\cite{ye2023mplug} &75.8 &68.8 &72.7 &68.4 &55.1 &62.1 &66.0 &65.0 \\
X-Decoder~\cite{zou2023generalized} &-- &-- &-- &-- &-- &-- &-- &64.6 \\
SEEM~\cite{zou2023segment}  &-- &-- &-- &-- &-- &-- &-- &65.7 \\
LISA-7B~\cite{lai2024lisa}  &76.5 &\underline{71.1} &74.1  &67.5 &56.5 &62.4 &68.5 &66.4 \\
PixelLM-7B~\cite{ren2024pixellm}   &76.5 &68.2&73.0  &\textbf{71.7} &\underline{58.3}&\textbf{66.3}  &\underline{70.5} &69.3 \\
OMG-LLaVA-7B~\cite{zhang2024omg}    & \underline{77.7} &\textbf{71.2} &\textbf{75.6} &69.7 &\textbf{58.9} &\underline{65.6} &70.2 &\underline{70.7} \\
\rowcolor{lightgray}
\textbf{Seg-R1-3B}$\diamond$  & 65.8 & 54.7& 58.7   & 56.2 & 45.0 & 49.1 & 57.0& 57.9 \\
\rowcolor{lightgray}\textbf{Seg-R1-3B}  & 76.0 & 64.9 & 69.9  & 66.8 &50.9 & 59.1 & 67.9 & 67.3 \\
\rowcolor{lightgray}\textbf{Seg-R1-7B}  & \textbf{78.7} & 67.6 & \underline{74.3}  & \underline{70.9} &57.9 & 62.6 & \textbf{71.4} & \textbf{71.0} \\
\bottomrule
\label{tab:refseg}

\end{tabular}
\end{minipage}
\hfill
\begin{minipage}[t]{0.41\linewidth}
\centering
\setlength{\tabcolsep}{1.0mm}
\scriptsize
\caption{Results on ReasonSeg.}
\label{tab:reasonseg}
\vspace{0.05in}
\begin{tabular}{c*{2}{c}*{2}{c}}
\toprule
& \multicolumn{2}{c}{val} 
& \multicolumn{2}{c}{test}  \\
\cmidrule(lr){2-3}\cmidrule(lr){4-5}
& gIoU  & cIoU
& gIoU  & cIoU \\
\midrule
OVSeg~\cite{liang2023open} &28.5 &18.6 &26.1 &20.8 \\
GRES~\cite{liu2023gres} &22.4 &19.9 &21.3 &22.0 \\
X-Decoder~\cite{zou2023generalized} &22.6 &17.9 &21.7 &16.3 \\
SEEM~\cite{zou2023segment} & 25.5 &21.2 &24.3 &18.7 \\
Grounded-SAM~\cite{liu2024grounding} & 26.0 &14.5 &21.3 &16.4 \\
LISA-7B~\cite{lai2024lisa} &44.4 &\underline{46.0} &36.8 &34.1 \\
\rowcolor{lightgray}
\textbf{Seg-R1-3B}$\diamond$ &50.3 &35.5 &42.4 &30.0 \\
\rowcolor{lightgray}\textbf{Seg-R1-3B} & \textbf{60.8} & \textbf{56.2}  & \underline{55.3} & \underline{46.6} \\
\rowcolor{lightgray}\textbf{Seg-R1-7B} & \underline{58.6} & 41.2  & \textbf{56.7} & \textbf{53.7} \\
\bottomrule
\end{tabular}
\end{minipage}
\end{table*}

\noindent\textbf{Zero-Shot Transfer.} Since reinforcement learning enables the model to acquire human-like prompting strategies for segmenting target objects, we further investigate whether this capability generalizes to other tasks. Specifically, we explore referring segmentation and reasoning segmentation—two challenging benchmarks that demand both visual and language understanding. Referring segmentation requires identifying and segmenting specific objects based on natural language expressions, while reasoning segmentation involves interpreting complex, multi-step instructions or contextual logic to produce accurate segmentation masks.

Surprisingly, we find that our model, trained solely via reinforcement learning on \(7,040\) foreground segmentation image-mask pairs without any textual supervision, exhibits strong zero-shot generalization to both tasks. On RefCOCOg~\cite{yu2016modeling}, Seg-R1 achieves \(71.4\) and \(71.0\) on the validation and test sets, comparable with the performance of current SoTA models. On the test set of ReasonSeg~\cite{lai2024lisa}, it reaches \(56.7\) in gIoU, significantly outperforming the previous models like LISA-7B~\cite{lai2024lisa}.

\begin{figure*}[h]
\centering
\includegraphics[width=1.0\columnwidth]{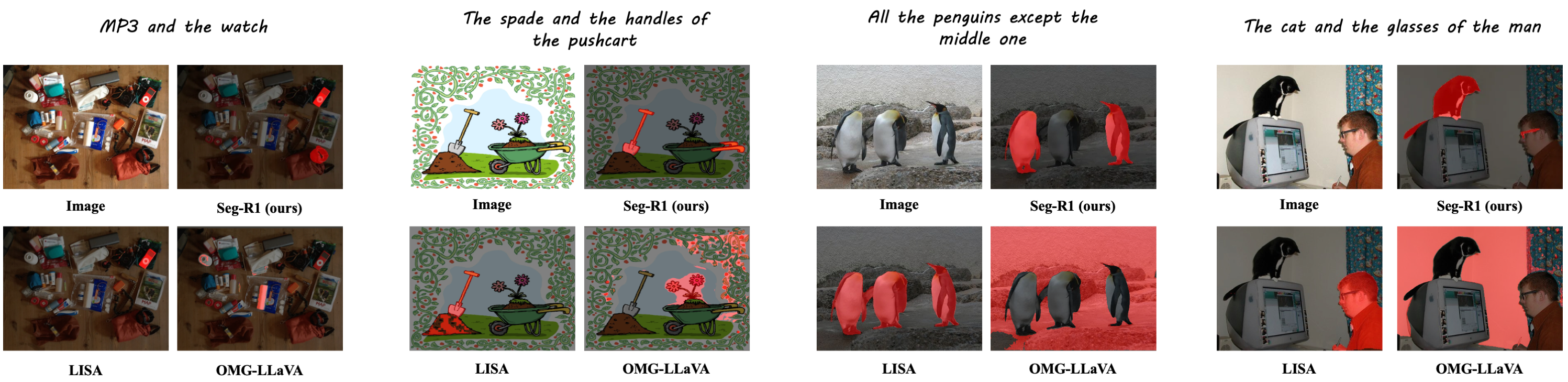}
\caption{Comparison of multiple objects referring segmentation in the wild.}
\label{fig:multiobject_comparison}
\end{figure*}

While previous models achieve high scores on benchmarks like RefCOCO, our evaluation reveals a significant gap between benchmark performance and real-world applicability. As shown in Figure~\ref{fig:comparison} and Figure~\ref{fig:multiobject_comparison}, we test the referring segmentation ability of the model in the wild and observe that existing models such as LISA~\cite{lai2024lisa} and OMG-LLaVA~\cite{zhang2024omg} fail on many cases involving real-world complexity, such as sketches and fine-grained segmentation. In contrast, Seg-R1 consistently demonstrates accurate segmentation across a wide range of open-vocabulary expressions. Whether in real-world images, hand-drawn illustrations, or multi-object scenarios, Seg-R1 shows remarkable generalization and robustness in understanding and grounding expressions for segmentation.

\subsection{Ablation Study}

\label{sec:ablation}


\begin{table*}[t]
\centering
\begin{minipage}[t]{0.4\linewidth}
\centering
\setlength{\tabcolsep}{1.0mm}
\small
\caption{Ablation on training strategy.}
\vspace{0.05in}
\begin{tabular}{l*{2}{c}*{2}{c}}

\toprule
 &$S_{\alpha}\uparrow$ & $F_{\beta}\uparrow$ & $E_{\phi}\uparrow$ & $M\downarrow$ \\
\midrule
baseline &.695 &.682 &.734 &.181\\
\(w.\) RL &.724 &.691 &.749 &.127 \\
\(w.\) SFT &.790 &.770 &.841 &.093\\
\rowcolor{lightgray}\(w.\) SFT + RL &.810 &.798 &.861 &.077 \\
\rowcolor{lightgray}\(w.\) Pre-RL + RL &.805 &.787 &.853 &.079 \\

\bottomrule
\end{tabular}
\label{tab:rl}
\end{minipage}
\hfill
\begin{minipage}[t]{0.56\linewidth}
\centering
\setlength{\tabcolsep}{1.0mm}
\caption{Ablation on segmentation reward function.}
\label{tab:reward_func}
\vspace{0.05in}
\small
\begin{tabular}{l*{2}{c}*{2}{c}}

\toprule
 &$S_{\alpha}\uparrow$ & $F_{\beta}\uparrow$ & $E_{\phi}\uparrow$ & $M\downarrow$ \\
\midrule
baseline &.614 &.532 &.439 &.151 \\
\(w.\) IoU &.785 &.770 &.889 &.084 \\
\(w.\) S-Measure &.410 &.049 &.251 &.181\\
\rowcolor{lightgray}\(w.\) IoU + S-Measure &.805 &.787 &.853 &.079 \\

\bottomrule
\end{tabular}
\end{minipage}
\end{table*}

\noindent\textbf{Effects of training strategy.} We compare the impact of different training strategies on model performance, as summarized in Table~\ref{tab:rl}. We use Qwen-2.5-VL-3B as our baseline and CAMO as the benchmark. Specifically, we prompt Qwen-2.5-VL to generate a bounding box for the target object, and then use it to guide SAM2 for mask generation. Our experiments show that both supervised fine-tuning (SFT) and reinforcement learning (RL) significantly enhance segmentation performance. Notably, SFT with the proposed FCoT dataset as cold start or pre-training with the high-resolution DIS5K dataset prior to RL both lead to substantial gains on the COD task. Building upon these improvements, further applying RL on COD10K brings additional benefits, consistently boosting segmentation performance. Besides, we observe a promising reward curve during the RL process as shown in Figure~\ref{fig:seg_reward}.

\noindent\textbf{Effects of Reward Function.}
We explore the impact of different reward functions during GRPO training. The model pre-RL on DIS5K serves as our baseline, and CAMO serves as the benchmark. As shown in Table~\ref{tab:reward_func}, using only IoU as the reward leads to significant performance improvements on the CAMO benchmark. However, relying solely on IoU may cause the model to overlook structural details within the mask. To address this, we also experimented with using only the S-Measure as the reward. The results show that this leads to reward hacking—without IoU as a corrective signal, the model tends to predict entirely black masks, resulting in poor performance on several metrics. To mitigate this issue, we adopt a combined reward function that incorporates both IoU and S-Measure. As demonstrated in the table, this combination consistently achieves the best results in COD tasks, effectively balancing overall mask accuracy with structural fidelity.

\noindent\textbf{Comparison of RL and SFT.} As shown in Tables~\ref{tab:cod} and~\ref{tab:sod}, pure RL strategy and SFT followed by RL yield comparable performance on foreground segmentation tasks. However, the pure RL strategy exhibits far superior generalization capabilities compared to the SFT-based method. As illustrated in Tables~\ref{tab:refseg} and~\ref{tab:reasonseg}, we evaluate both models on two out-of-domain tasks: referring segmentation and reasoning segmentation. We find that the RL-trained model successfully transfers its learned segmentation skills from foreground segmentation to these new domains by leveraging the strong visual reasoning and grounding capabilities of Qwen-2.5-VL. In contrast, the SFT-based model performs relatively poorly on both tasks.

\begin{table*}[t]
\begin{minipage}[t]{0.45\linewidth}
\centering
\figcaption{Visualization of segmentation reward curve.}
\vspace{0.05in}
\includegraphics[width=\linewidth]{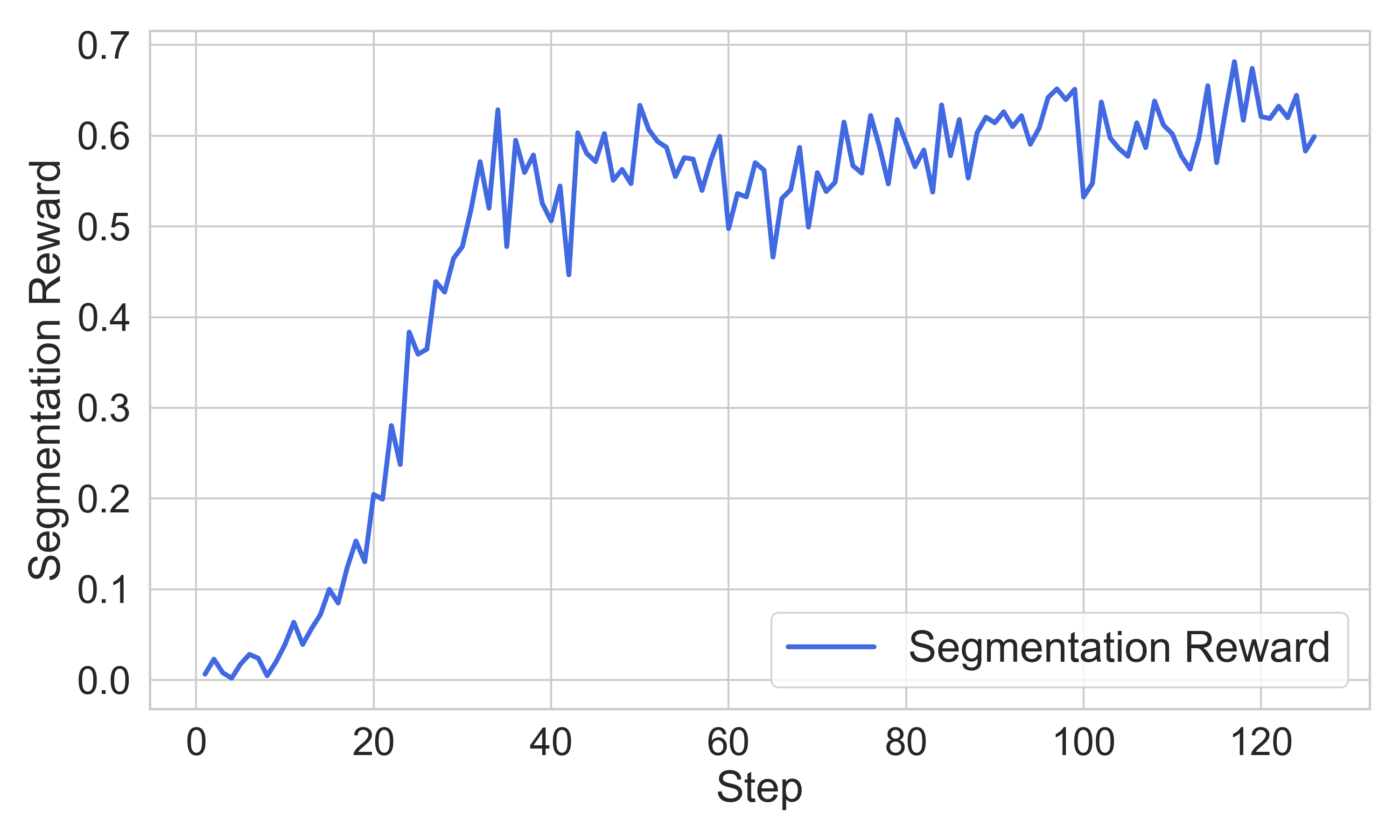}
\label{fig:seg_reward}
\end{minipage}
\hfill
\begin{minipage}[t]{0.52\linewidth}
\centering
\figcaption{Comparison of the general VLM benchmarks.}
\vspace{0.05in}
\includegraphics[width=\linewidth]{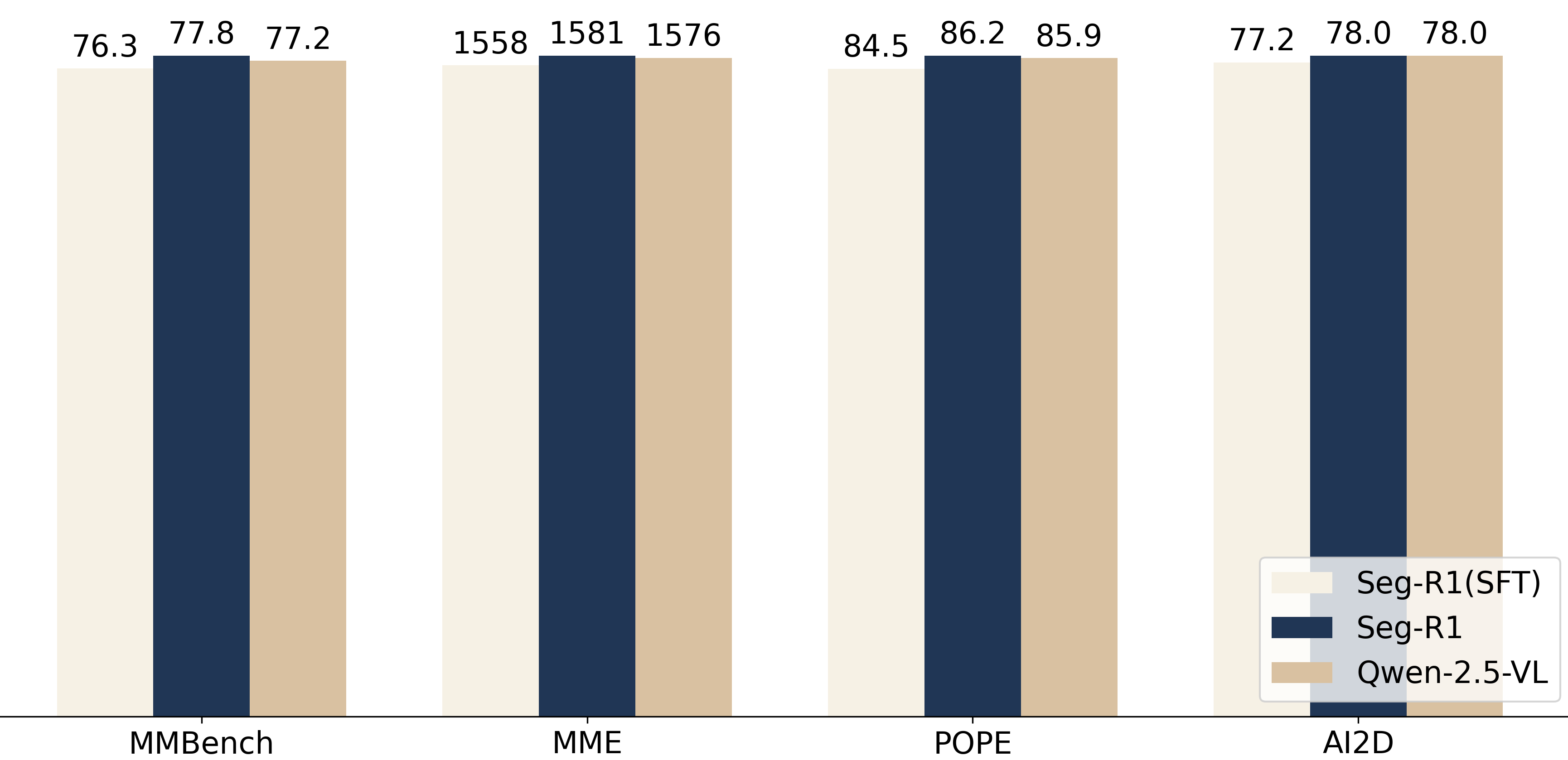}
\label{fig:general_tasks}
\end{minipage}

\end{table*}

Moreover, we further evaluate the models on a range of general multimodal benchmarks. As shown in Figure~\ref{fig:general_tasks}, the pure RL version of Seg-R1 retains performance on par with the original Qwen-2.5-VL across MMBench~\cite{liu2024mmbench}, MME~\cite{fu2023mme}, POPE~\cite{li2023evaluating}, and AI2D~\cite{kembhavi2016diagram}. In contrast, the SFT version suffers from performance degradation, indicating a loss of general capabilities.

These findings highlight a key advantage of the RL approach: it equips the model with strong segmentation capabilities without sacrificing its original reasoning and understanding abilities, whereas SFT tends to overfit to the segmentation task at the expense of general multimodal performance.


\section{Conclusion}

In this work, we propose \textbf{Seg-R1}, a simple yet highly effective paradigm for enabling large multimodal models (LMMs) to perform segmentation. Instead of relying on dense mask generation or complex architectural modifications, we reformulate segmentation as a next-mask-prompt prediction task, where the LMM generates a sequence of mask prompts that guide SAM2 to produce high-quality masks. This design significantly reduces the learning cost of segmentation and aligns naturally with the causal structure of autoregressive language models. Through a carefully designed RL framework, we demonstrate that: (1) Pure RL can endow LMMs with powerful segmentation abilities. Seg-R1 achieves state-of-the-art performance on multiple segmentation benchmarks; (2) Compared to SFT, RL offers significantly stronger generalization. Trained solely on \(7,040\) foreground segmentation image-mask pairs, Seg-R1 exhibits impressive zero-shot performance on both referring segmentation and reasoning segmentation tasks, even under open-world settings; (3) RL preserves the original general-purpose capabilities of the LMM, whereas SFT tends to compromise performance on general visual comprehension. 

Despite these promising results, there remains room for future exploration. For instance, further improve the performance of Seg-R1 on camouflaged object detection benchmarks and extend support to multi-turn interaction. We plan to explore these in the future.


\bibliographystyle{abbrv}
\bibliography{neurips_2025}

\end{document}